\documentclass[journal,12pt,onecolumn,draftclsnofoot]{IEEEtran}
\usepackage[T1]{fontenc}

\ifCLASSINFOpdf
\else
\fi
\usepackage{amsmath}
\usepackage{graphicx}      
\usepackage{stackengine}
\usepackage{url}
\usepackage{amsmath,amssymb}
\usepackage{graphicx}
\usepackage{epstopdf,subfigure}

\newcommand{\vbar}{\raisebox{.17ex}{\rule{.04em}{1.35ex}}}
\newcommand{\vbarind}{\raisebox{.01ex}{\rule{.04em}{1.1ex}}}
\newcommand{\D}{\ifmmode {\rm I}\hspace{-.2em}{\rm D} \else ${\rm I}\hspace{-.2em}{\rm D}$ \fi}
\newcommand{\T}{\ifmmode {\rm I}\hspace{-.2em}{\rm T} \else ${\rm I}\hspace{-.2em}{\rm T}$ \fi}
\newcommand{\B}{\ifmmode {\rm I}\hspace{-.2em}{\rm B} \else \mbox{${\rm I}\hspace{-.2em}{\rm B}$} \fi}
\newcommand{\Hil}{\ifmmode {\rm I}\hspace{-.2em}{\rm H} \else \mbox{${\rm I}\hspace{-.2em}{\rm H}$} \fi}
\newcommand{\C}{\ifmmode \hspace{.2em}\vbar\hspace{-.31em}{\rm C} \else \mbox{$\hspace{.2em}\vbar\hspace{-.31em}{\rm C}$} \fi}
\newcommand{\Cind}{\ifmmode \hspace{.2em}\vbarind\hspace{-.25em}{\rm C} \else \mbox{$\hspace{.2em}\vbarind\hspace{-.25em}{\rm C}$} \fi}
\newcommand{\Q}{\ifmmode \hspace{.2em}\vbar\hspace{-.31em}{\rm Q} \else \mbox{$\hspace{.2em}\vbar\hspace{-.31em}{\rm Q}$} \fi}
\newcommand{\Z}{\ifmmode {\rm Z}\hspace{-.28em}{\rm Z} \else ${\rm Z}\hspace{-.38em}{\rm Z}$ \fi}

\renewcommand{\vec}[1]{{\bf{#1}}}     

\renewcommand{\B}{_{\rm{B}}}
\begin{document}
%
\title{The Internet of Things as a Deep Neural Network}
%
%
%

\author{Rong~Du\thanks{The paper is under review in IEEE Communications Magazine}, Sindri~Magn\'usson, and~Carlo~Fischione}

\maketitle

\begin{abstract}
An important task in the Internet of Things (IoT) is field monitoring, where multiple IoT nodes take measurements and communicate them to the base station or the cloud for processing, inference, and analysis. 
This communication becomes costly when the measurements are high-dimensional (e.g., videos or time-series data). The IoT networks with limited bandwidth and low power devices may not be able to support such frequent transmissions with high data rates.
To ensure communication efficiency, this article proposes to model the measurement compression at IoT nodes and the inference at the base station or cloud as a deep neural network (DNN). We propose a new framework 
where the data to be transmitted from nodes are the intermediate outputs of a layer of the DNN. We show how to learn the model parameters of the DNN and study the trade-off between the communication rate and the inference accuracy. The experimental results show that we can save approximately 96\% transmissions with only a degradation of 2.5\% in inference accuracy. Our findings have the potentiality to enable many new IoT data analysis applications generating large amount of measurements.
\end{abstract}


%
\IEEEpeerreviewmaketitle

\section{Introduction}
%
%
%
%
Distributed detection, monitoring, and classification are
important tasks in wireless sensor networks or the Internet of Things. In these tasks, sensor nodes collect measurements
and send them to a base station/gateway, and further to an application server in the IP network that can then
make a decision or an inference based on the measurements from the nodes. These applications include camera surveillance to detect intrusions or suspicious events, environmental monitoring to detect and track pollutions, and structural health monitoring to prevent structural failures.

If the inference  has to be done frequently and the
data is high dimensional, then the nodes may need to transmit large amount of data to the base stations, which is costly in bandwidth and may be infeasible with the existing IoT networks. Consider the communication standard Narrow Band IoT, which is a
cellular based IoT standard whose uplink physical layer
throughput is 200-250~kbps. If a cellular base station supports 20000 nodes, which is
around half of its maximum capacity, then the 
average rate per node is less than 10-12.5~bps. Moreover, it is inefficient to let the nodes communicate the full measurements, since they typically have redundant and correlated information. Unless we reduce the transmitted data, the number of nodes that can be served and their data rates will be greatly limited.

To reduce the amount of data to transmit, there are some classic methods that could be used for IoT systems, such as distributed source coding~\cite{xiong2004distributed}, distributed estimation~\cite{xiao2006distributed}, and distributed detection~\cite{chamberland2007wireless}. However, these methods usually require a good knowledge about the underlying correlation model among the measurements taken at different sensors. For example, distributed source coding and distributed detection rely on the correlation of the nodes, and distributed estimation requires knowledge about the model of the measurements and the system state, which may be not easy to get in complex systems.

\begin{figure*}[t]
	\centering
	\subfigure[distributed datasets]{\includegraphics[height=1.65in]{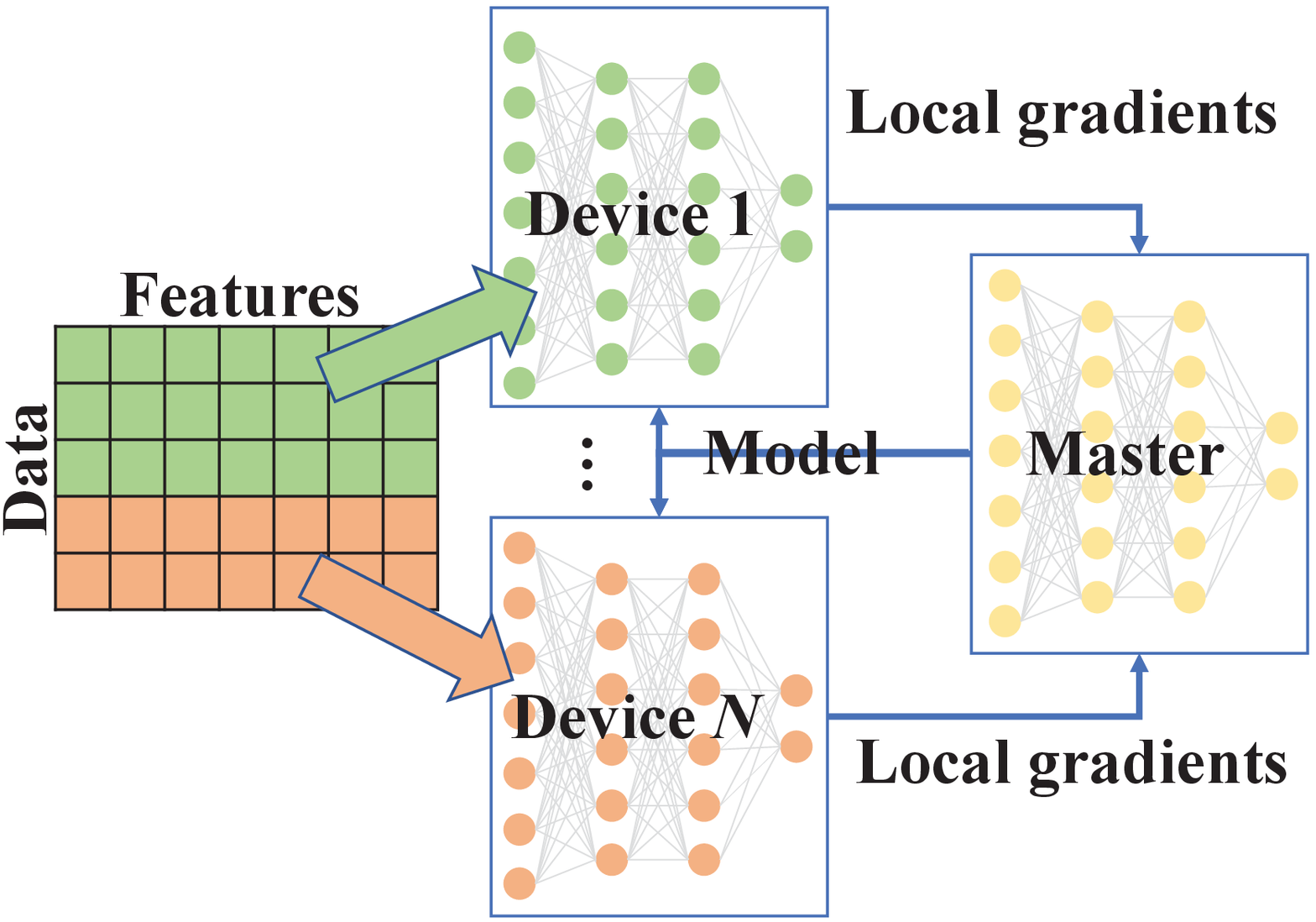}}\hspace{0.2in}
	\subfigure[distributed features]{\includegraphics[height=1.65in]{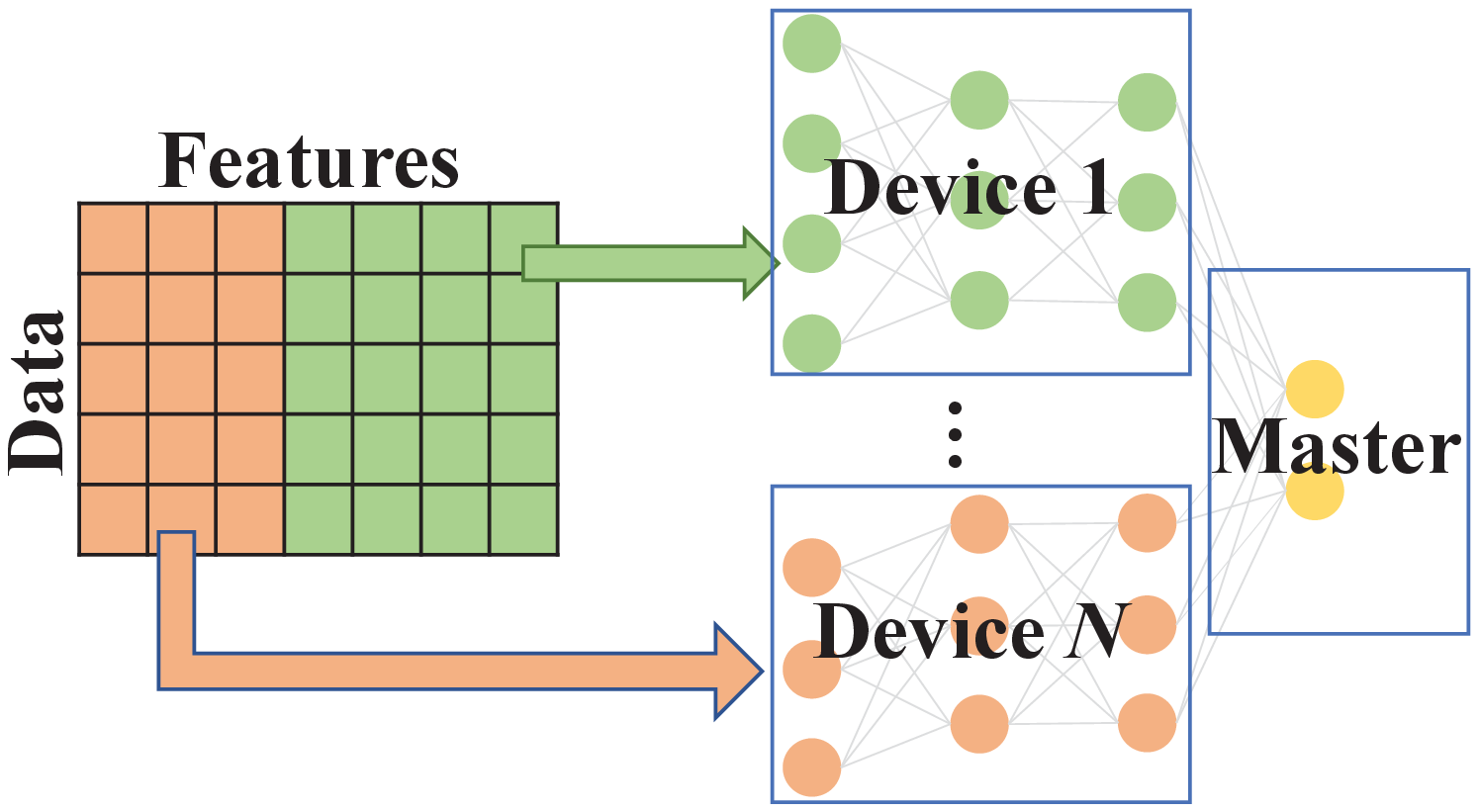}}\hspace{0.2in}
	\subfigure[distributed models]{\includegraphics[height=1.65in]{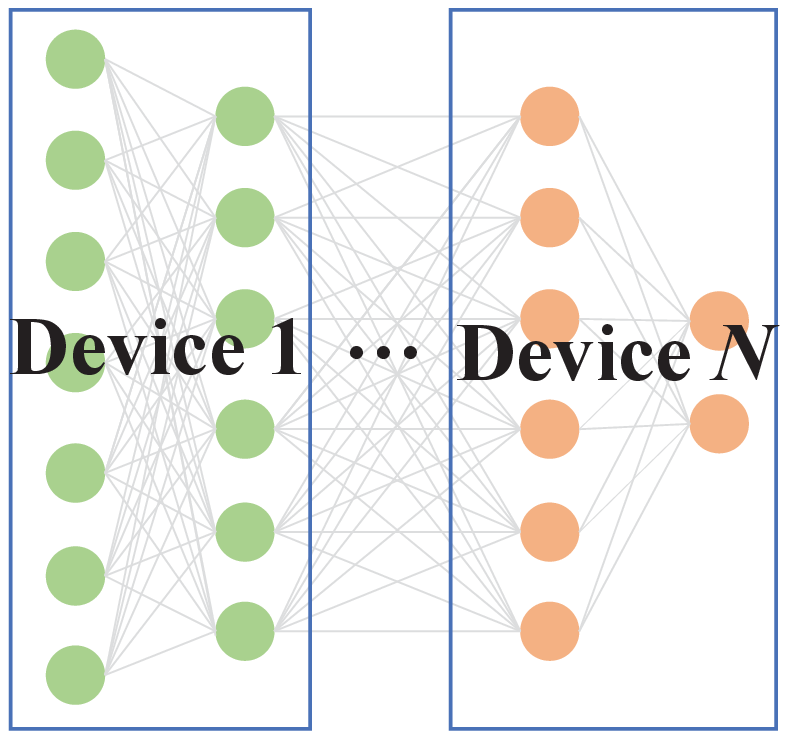}}
	\vspace{-3mm}
	\caption{The three popular frameworks to distribute the ML to multiple devices, according to the state of the art~\cite{konevcny2016federated,hu2019fdml,teerapittayanon2017distributed}.}
	\label{fig:DistriL}
	\vspace{-4mm}
\end{figure*}

The alternative to the model based approach mentioned above is Deep Learning (DL), which has become the state of the art in machine learning (ML) and is widely used in different monitoring applications with IoT~\cite{abeshu2018deep,wang2018rf}. However, since sending large amount of high dimensional raw data from massive IoT nodes will congest the IoT network, the collect-then-infer scheme of DL does not scale well with increased numbers nodes and applications of in IoT systems (massive IoT). To handle the scalability issue, distributed learning is a promising solution. There are two challenges for using distributed ML in IoT. One is the limited bandwidth in IoT networks, which requires us to reduce the data transmission in both training and inference phases. The other one is the limited computation, storage, and energy capacity of IoT nodes, which means that the ML model at the IoT nodes should be tailored according to their capacity. Edge computing provides gateways and base stations with storage and computing capacity, and it enables us to allocate some inference processes to the edge devices (data gateways/base stations). The additional energy consumption of the IoT nodes in computing could be compensated if they can save energy from sending less data. Notice that the DL architecture, deep neural network (DNN), is a network for inference with multiple layers, and an IoT network is a network for communication also with a hierarchical architecture (end devices, edge devices, and cloud. To make the paper compatible with both cloud and edge computing architecture, we consider all devices that are not IoT nodes as a whole and call it on the base station side.). The question is whether it is possible to adapt the DNN layers and their connections to the physical IoT network while maintaining a good DL performance. 

This article investigates the question above, and we propose the idea of modelling IoT networks as DNN to reduce the data transmission for inference. Different to the traditional DL, we consider the cost in communications in terms of the amount of transmitted data, due to the limited bandwidth of IoT networks, especially from the nodes to the edge, which is helpful to reduce delay and energy consumption. We suggest to perform DL distributedly where the design of a DNN is seen as an abstraction of the IoT network capable to use less data transmissions while offering a good inference performance. We propose to design compression functions for each node that takes the high dimensional measurements as input and returns low dimensional representation of those measurements, and design inference function for the base station and upper layers in the communication network to aggregate the low dimensional information from all nodes to make the final DL prediction. In such a way, the compression function on each IoT node, as well as the inference function on the base station, becomes a DNN. Using the idea of ML over networks, we co-train all the compression functions and the inference algorithm together, to enforce the compression to extract the most useful information for the inference.


\section{Overview of DL Frameworks}

There are three main ways of distributing the ML tasks in networks with many nodes (see Fig.~\ref{fig:DistriL}): (a) to distribute the data samples between different nodes, (b) to distribute different features of the data to different nodes, and (c) to distribute the model among the nodes. We now review these approaches. In Fig.~\ref{fig:DistriL}, for illustration purpose we represent the data as matrix where rows are samples and columns the features. The main difference among the three frameworks is how a DNN is learned by the nodes distributedly.


\subsection{Learning with distributed datasets}
ML with distributed datasets is to learn a unified model with different nodes, each of which has local data samples sharing the same features space. For example, smart phones can learn and improve the ML model for image classification based on their local albums, and exchange some information about the model with each other to improve the accuracy of the model. It saves the transmission of getting the raw datasets from different nodes, especially in the training phase. Thus, this framework suits for the monitoring applications where the inference only needs the data at one node, and different nodes use a universal model.
%
%
%
%

The most common distributed framework is with
worker nodes and a master. A promising instance of the framework is Federated Learning~\cite{konevcny2016federated,mcmahan2016communication} that aims at achieving a universal ML model for all workers in a collaborated manner without the need of globally collecting raw data from the workers. In FL, each worker node downloads a global model and updates it based on local data before uploading it back to the master node. Then the master node averages the local models to update the global model. Since each node has a complete	 ML model locally, a node only needs to feed its measurements to the ML model to make an inference. 
Similar algorithmic ideas have been used for fully distributed ML (without a master node) where the worker nodes coordinate over communication networks with arbitrary topology (see~\cite{nedic2018network} and references therein). We can see that different entities only share the model parameters. Since the learning on distributed dataset requires the nodes to have data instances with a same feature space, it does not suit large scale monitoring that uses multiple IoT nodes to collect the data of different features.



\subsection{Learning with distributed features}
Learning with distributed features is fundamentally different to FL. It naturally appears when multiple nodes observe different features. For example, in surveillance camera networks, each camera captures the video providing information from different angles as different features. In large scale environmental monitoring such as fresh water monitoring, the measurements of various types of sensors at different locations give different features. The naive way to handle the distributed features is to collect all the data on a single node that can then train the ML model centrally. However, this is inefficient when the size of the raw data at each node is large. For such problems, distribute learning algorithms are needed.

Distributed learning algorithms that allow decomposing the  features among nodes have been considered in the literature, such as the ones based on dynamic diffusion approach~\cite{ying2018supervised} and adaptive direction multiplier method~\cite{hu2019learning}. In these schemes, the nodes calculate the inner product of the data and the model, and share the results with others. From the privacy point of view, the work in~\cite{hu2019fdml} proposed a feature distributed machine learning (FDML) framework, where each data owner proceeds the ML locally using local data and transmits the local prediction to a master node. The master node calculates the weighted sum of these local predictions and passes it to an activation function to make the final inference. In comparison, our scheme have multiple layers at the master node to achieve a better inference accuracy. In addition, the goal of the worker nodes in our scheme is to compress the information rather than making a local inference.

\subsection{Learning distributed models}
This framework does not relate to how data is distributed, but how the model is distributed. Instead of having the entire ML model at a single node, the framework distributes the ML model to multiple nodes. More specifically, a node is responsible to learn a subset of model parameters. Since DNN contains multiple layers, the most common way to distribute the model parameters is to partition the DNN according to the layers. A node with better storage and computation capability could be assigned with more layers. The inference is done sequentially from one node to another, by sharing intermediate results, till the last node where final inference is done.

The system in~\cite{teerapittayanon2017distributed} maps a DNN onto heterogeneous nodes. Each node represents some layers of the DNN and performs a local parameter update during training. This distributes the computational workload to multiple nodes to achieve a faster training and inference. However, such an approach is not thought for the IoT, where communication efficiency is a must, whereas the many links among the layers of this approach may demand huge bandwidth. In addition, it is not applicable when different nodes measure data with different features, unless a gateway gather all raw measurements from the IoT nodes.

Our framework uses the idea of learning with distributed features and the learning of distributed models, but we propose to greatly improve the communication efficiency of the ML in IoT networks. Briefly speaking, on the IoT nodes side, it is a feature distributed framework such that each IoT node makes some preliminary compression and inference based on its local measurement. From the system scale, the learning model is distributed among IoT nodes, gateways, and cloud to better handle different storage and computation resource at different nodes. Thus, it is also a ML with distributed models. We describe the details in the next section.




\section{Proposed Communication Efficient ML for IoT}
To enable ML to scale with massive IoT, we need a new framework/architecture of ML that is communication efficient. In this section, we will introduce our proposed framework.
\begin{figure*}[t]
	\centering
	\includegraphics[width=0.7\textwidth]{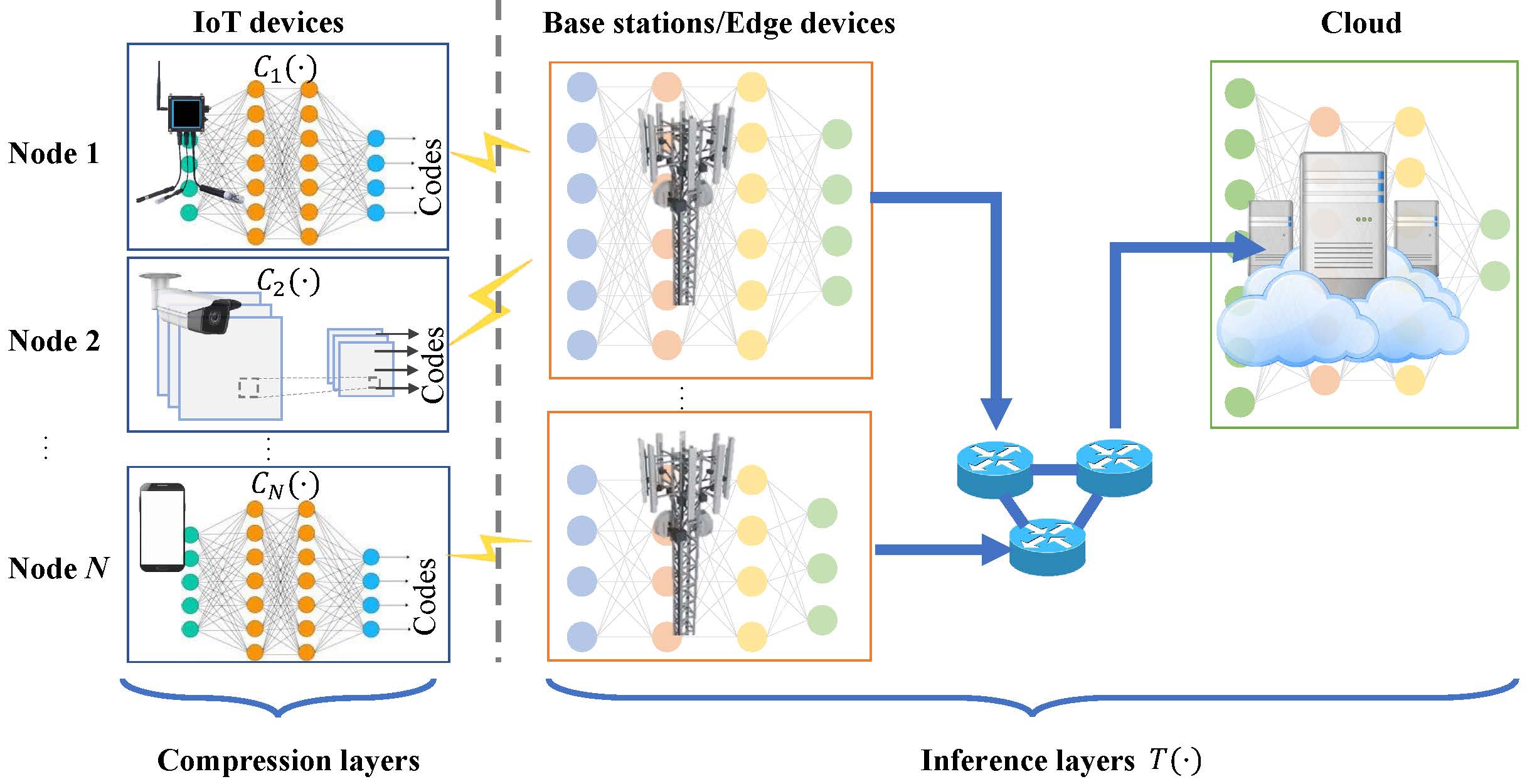}
	\vspace{-3mm}
	\caption{Our proposed ML framework for IoT that adapts a DNN to the distributed architecture of the IoT nodes, edge devices, and the cloud. The nodes, base station, and the cloud use deep neural networks to perform the compression and inference task, where the model parameters of the deep neural networks should be learned. For the sake of communication efficiency, the output from the nodes must have as little bits as possible while it should not harm the inference performance at the base station/cloud.}
	\label{fig:sysstruct}
	\vspace{-4mm}
\end{figure*}
\subsection{System architecture}
Consider an IoT network of $N$ remote IoT nodes that connect to a cloud application server through base stations, which is a typical cellular IoT structure~\cite{zayas20173gpp}. 
The goal of the system is to perform an inference based on a feature vector $\vec{x}$ with dimension $d$, which consists of the feature vector $\vec{x}_i$ with dimension $d_i$ that is captured by node $i$. The nodes are used for a monitoring task that detects the event representing one of the possible classes. ML is applied to find an inference rule, such that given an input data measured by the nodes as generated by the event, the inference result is the true class that the event belongs to. 

To perform the inference, each node communicates some information $\vec{c}_i$ to the base station. We could set $\vec{c}_i=\vec{x}_i$ meaning that each node communicates its full feature vector. However, this is inefficient. 
To make the communication efficient, the size of $\vec{c}_i$ should be as small as possible. 

We let $\vec{c}_i=C_i(\vec{x}_i)$ be a compressed version of $\vec{x}_i$, where $C_i(\cdot)$ is a compress function to be designed. Note that, we use compression to represent the process that reduces the data size from the raw data to the transmitted information from a node. The process could include extraction, summarization, coding, etc. Such a compression process can be done when the IoT node is waiting for its transmission window. Having gathered $\vec{c}_i$ from all nodes as $\vec{c}$, the base stations and the cloud perform the inference $T(\vec{c})$. Since the compression process reduces the data to be transmitted, it will reduce the communication delay and thus reduce the time of the entire inference process. The main challenge in designing this communication-efficient ML is to design appropriate compression functions $C_i(\cdot)$ and the inference function $T(\cdot)$. We next illustrate how we can build them as DNNs and get the function coefficients by training the DNNs.


\subsection{IoT as DNN}
Notice that both IoT communication network and DNN inference network have a structure of multiple layers. Thus, it is natural to see IoT networks as a DNN, where the IoT nodes correspond to the shallow layers of the entire DNN and are responsible for extracting the information to lower the data to transmit; whereas the base stations and cloud correspond to the deep layers of the DNN and are responsible for making inference based on the information sent from the IoT nodes. As a result, we can jointly learn the compression functions $C_i(\cdot)$ and inference function $T(\cdot)$ by ML. We design $C_i(\cdot)$ and $T(\cdot)$ as the layers structure of a DNN,  such as the types of the layers and the number of neurons at each layer. Then, the compression functions and inference function become the model parameters of the DNN to be learned. For example, a fully connected layer corresponds to a function $\sigma(\vec{W}\vec{x}_{\textrm{in}}+\vec{b})$, where $\vec{x}_{\textrm{in}}$ is the input of this layer and is the output of its previous layer, $\vec{W}$ and $\vec{b}$ are the model parameter to be learned of the layer, and $\sigma(\cdot)$ is an activation function, e.g. a sigmoid function. When $\vec{W}$ is a fat matrix, the output has a smaller dimension than the input and thus the layer corresponds to a compression; otherwise, the layer projects a low dimensional input to a higher dimensional vector and can be used for inference. Once the model parameters of each layer in the DNN have been trained, as will be discussed in the Section~III.C, the compression functions and inference function are the compositions of multiple functions, each of which corresponds to a layer in the DNN.

As depicted in Fig.~\ref{fig:sysstruct}, the entire system can be considered as a DNN, where the nodes correspond to the shallow disjoint parts of the DNN, and the base station and cloud corresponds to the deep part. Using the training data collected by the IoT nodes, then we can find the optimal functions $C_i(\cdot)$ and $T(\cdot)$  by minimizing a loss function that captures the inference accuracy and the complexity of the functions $C_i(\cdot)$ and $T(\cdot)$. Once the model parameters of the DNN are trained, we can achieve the compression and inference functions by the stacking of the layers in the DNN. Typical examples of the layers include convolutional layers, pooling layers, fully connected layers, or recurrent layers. Differential privacy can be used at the output to ensure data privacy during the inference stage. The idea of summarization~\cite{muhammad2019cost} could be used to discard unimportant data. Quantization can also be applied at the output of the nodes to reduce the bits to transmit. Since we aim for a communication efficient framework, the nodes should reduce the size of their outputs to compress the information, and the base station should expand the inputs at the first hidden layer to extract the compressed information. Our proposed scheme is illustrated in Fig.~\ref{fig:sysstruct}.

The following three schemes are special cases of our proposed scheme:
\begin{itemize}
	\item Direct transmission of raw data: It means that the only learning is at the base station part, using the global data. Thus, it might have the best learning performance, but the communication cost is high and not accord with the nowadays massive IoT concept.
	\item Autoencoder with inference: Autoencoder is an artificial neural network that learns data coding to reduce the dimension of the data. The nodes can first use an autoencoder to compress the sensory data to codes, and the base station reconstructs the data from the codes. Afterwards, the base station applies a standard DNN for inference. However, consider that the goal of the monitoring application is to have a good inference accuracy, instead of reconstruction accuracy. It is not necessary to have a reconstruction part at the base station, which makes the entire DNN more complex. To be more efficient, our framework directly makes inference based on the encoded information, which can be seen as to directly combine the reconstruction layers and inference layers at the base station side in this autoencoder with inference scheme.
	\item Feature distributed ML in~\cite{hu2019fdml}: It uses a shallow neural network with no hidden layer at the base station and the output from each node is just a real value.  This scheme might save a lot in the communication from nodes to the base station. However, since the model at the base station side might be too simple, the degradation of learning performance may be high, as we will show in the simulations. 
\end{itemize}
In our framework, since we focus on communication efficiency, we do not require the DNNs at the node side or the base station side to be shallow. Instead, we require that the output from the nodes should be small without harming the inference performance. 

The DNN at the base station side could be further partitioned into two parts, as shown in Fig.~\ref{fig:sysstruct}: one as the DNNs at some edge devices, and the other as the DNN at the cloud center, such that the framework could support the applications that contain nodes in multiple cells. However, in such cases, since the wireless communication from the edge devices to the cloud usually has a large bandwidth, the bottleneck of the communication is more likely between the nodes and the edge devices. Therefore, we mainly focus on the data that are sent by the nodes.

\subsection{Training of the model parameters}
Since we design the compression functions and the inference functions as the layers of a DNN, the coefficients of these functions are the model parameters of the DNN. To achieve the coefficients, we can use a centralized method. More specifically, a centralized cloud center collects all training data and the corresponding desired output. Then, it trains the parameters of the entire DNN in a centralized manner and sends back the models to the nodes and the base station. The IoT nodes and the base station then apply the trained model during inference.

More specifically, denote by $\vec{\bar{x}}_i\in \mathbb{R}^{d_i\times m}$ the training data of IoT node $i$, where $m$ is the sample size. Then, the training data samples of the system $\vec{\bar{x}}\in\mathbb{R}^{d\times m}$ is achieved by stacking $\vec{\bar{x}}_i$ for all $i$. We denote $\vec{\bar{y}}$ the corresponding output. Let $\vec{w}=\{\vec{w}_0,\ldots,\vec{w}_N\}$ be the ML model parameters of the entire DNN, where $\vec{w}_0$ is the model parameters of $T(\cdot)$, and $\vec{w}_i, (i=1,\ldots,N)$ is the model parameters of $C_i(\cdot)$. Then, the loss function for the training can be written as
\begin{align*}
&L(\vec{\bar{x}},\vec{\bar{y}};\vec{w})=\sum_{i=0}^N R_i(\vec{w}_i)+D\left(T(C_1(\vec{\bar{x}}_1;\vec{w}_1),\ldots,C_N(\vec{\bar{x}}_N;\vec{w}_N);\vec{w}_0),\vec{\bar{y}}\right),
\end{align*}
where the first term $R(\cdot)$ is the regularization that corresponds to the complexity of the ML model, and the second term $D(\cdot,\cdot)$ is the loss term that corresponds to the accuracy of the inference. The model $\vec{w}$ is learned by minimizing $L(\vec{\bar{x}},\vec{\bar{y}};\vec{w})$, which can be done by iteratively updating
\begin{align*}
\vec{w}\leftarrow \vec{w}-\eta \nabla_{\vec{w}} L(\vec{\bar{x}},\vec{\bar{y}};\vec{w})\,,
\end{align*}
where $\eta$ is a predefined step size of learning. When the training fishes, the central cloud sends $\vec{w}_i$ to IoT node $i$. It only requires the IoT nodes to have sufficient storage capacity to store the model and the computational capacity to calculate $\vec{c}_i$ according to its measurements and trained model.


We could also train the DNN in a distributed manner, i.e., the nodes and the base station train their local parameters locally. More specifically, the nodes and the base station first initialize their local parameters. With the model parameters and a batch of training data, each node calculates its codes as the output of its compression function and sends the codes $\vec{c}_i=C_i(\vec{\bar{x}};\vec{w}_i)$ to the base station. Using the received codes from all nodes as the input and the desired output $\vec{\bar{y}}$, the base station calculates the gradients 
\begin{align*}
\nabla_{\vec{w}_0} L=\left.\frac{\partial D}{\partial T}\frac{\partial T}{\partial \vec{w}_0}\right|_{\vec{c}_1,\ldots,\vec{c}_N,\vec{\bar{y}}}+\frac{\partial R_0}{\partial \vec{w}_0}
\end{align*}
and makes an update of the parameters in the inference function as $\vec{w}_0\leftarrow \vec{w}_0-\eta \nabla_{\vec{w}_0}L$. Then, it sends the gradients to the nodes. Based on the received gradients from the base station, each node then calculates the gradients with respect to its local parameters
\begin{align*}
\nabla_{\vec{w}_i} L=\left.\frac{\partial L}{\partial C_i}\right|_{\vec{w}_0,\vec{\bar{y}},\vec{c}_{j} (j\neq i)}\left.\frac{\partial C_i}{\partial \vec{w}_i}\right|_{\vec{\bar{x}}_i}+\frac{\partial R_i}{\partial \vec{w}_i}\,,
\end{align*} 
where $\partial L/\partial C_i|_{\vec{w}_0}$ is calculated and sent from the base station. Then, the IoT node makes an update of the parameters by $\vec{w}_i\leftarrow \vec{w}_i-\eta \nabla_{\vec{w}_i}L$. Such an update of parameters process continues until convergence. During the training process, the nodes transmit neither the raw training data nor their local model parameters, which allow them to preserve the data privacy. This distributed scheme requires the IoT nodes to have sufficient computational power to calculate the local gradients and storage capacity to store the training data set.  If the IoT nodes' local computation and storage capacity are not sufficient for the training, they can offload the task and their training data to a trusted device, such as a private gateway or server, and get the local ML model when the training finishes.

\section{Application and Numerical Results} 
In this section, we will study the performance of the proposed distributed ML framework by simulations. We assume that the data transmission is ideal, i.e., the received and decoded data is the same as the transmitted data, which can be achieved using error correction coding and retransmission. We first examine our method by using the MNIST data set, and then with a real-world data set from a smart water monitoring system.

\subsection{MNIST data set}
The classical data set MNIST is a large dataset of handwriting digits. The input feature vector is an image with $28\times 28$ bytes, and the labels are one of the $10$ digits.

\begin{figure}[t]
	\centering
	\includegraphics[width=0.45\textwidth]{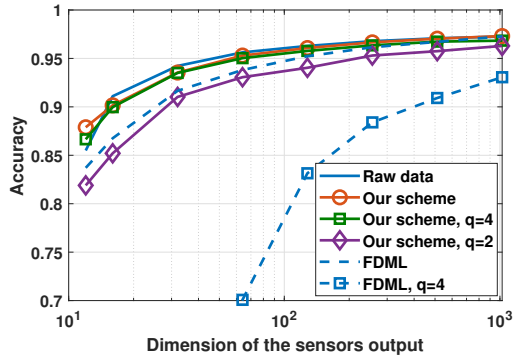}	
	\vspace{-4mm}
	\caption{Comparison of the classification accuracy achieved by our scheme and the FDML scheme with different quantizations. }\label{fig:multiple_comp_vs_acc}
	\vspace{-5mm}
\end{figure}

We consider the cases with four IoT nodes. We assume that the first, second, third, and the fourth node capture the top-left, top-right, bottom-left, and bottom-right of the figures, respectively, i.e., each of them will have $14\times 14$ bytes input. For each node, its first compression layer is a convolution layer with 32 $5\times5$ convolution filters, and a max pooling of $2\times 2$, which gives $7\times 7\times 32$ outputs. The next hidden layer is a compression layer (a fully connected layer with $C/4$ outputs), where $C$ is the output dimension of the nodes to the base station. Quantization could be applied on the output data to further compress the data for transmission. Each node transmits the quantized outputs to the base station. If $q$-bits quantization is used, the number of bits that a node sends is $qC/4$, and the base station will receive $C$ $q$-bits. The base station feeds these $C$ $q$-bits data into a fully connected layer with $1024$ outputs. The output layer is a softmax layer with 10 outputs. The result is shown in Fig.~\ref{fig:multiple_comp_vs_acc}. The X axis is $C$, which is taken to be 12, 16, 32, 64, 128, 256, 512, and 1024, respectively. We compare the classification accuracy in the cases of no quantization and quantization with $q$ be 1, 2, and 4 at the nodes outputs. We also use as a benchmark to mimic the case where all nodes transmit raw data, and the learning is done based on entire features. 

In the cases with multiple nodes, each node only utilizes local observation to learn locally and distributedly, and thus the prediction accuracy is lower than that of the centralized scheme. However, the difference is not large, which suggests that the loss due to local learning on distributed features is tolerable for the sake of communication and privacy. When we compare the performances of different quantizations, we observe that, with 4-bits quantization (the green line with squared marks), the performance is similar to the case without quantization (the red line with circle marks). It suggests that 4-bits quantization is good enough in reducing the data to be transmitted from the nodes. Comparing the case where raw image is sent, which requires $28\times 28\times 8$ bits, the scheme having 64 outputs in total and using 4-bits quantization only needs to send $256$ bits for each inference. Although the accuracy drops by approximately $2.5\%$, our framework saves 96~\% in data transmission.

Last, we compare the performance of our framework with the FDML framework~\cite{hu2019fdml}. At the nodes side, the setups of the DNN for FDML and our framework are the same. At the base station side, the FDML approach calculates the weighted sum of the received data and passes the result to an activation function and a softmax function. We also test the classification accuracy with 4-bits quantization before data transmission. The result is shown in the dashed lines of Fig.~\ref{fig:multiple_comp_vs_acc}, where the line with squared marks means the case with 4-bits quantization. When the nodes do not apply quantization, we observe that the accuracy achieved by our scheme is higher than the FDML scheme. The improvement is larger when $C$ is small. Such a gain comes from the non-simple learning model that is applied at the base station side, instead of being a simple weighted sum and activation approach in FDML. Such a gain is significantly larger when the quantization is applied. In our scheme, the difference in accuracy with and without quantization is only around $1\%$, which is almost negligible. However, for the FDML scheme, the difference is large, especially when $C$ is small. It indicates that, when we have compression at the nodes side, especially quantization is also applied, to reduce the bits for transmission from the nodes, it is important to apply expansion at the base station side to extract the compressed information, such that the compression in data transmission will not harm the learning performance.


\subsection{iWater Data set}

Next, we test the framework using the real-world sensory data from Vinnova project iWater~\cite{iwater}. The project uses sensors to monitor the water conditions in Lake M\"{a}laren, Sweden. We focus on the Optical Dissolved Oxygen level (ODO) data from 4 different locations during the period from Oct. 26th, 2016 to Apr. 21st, 2017. We build a DNN based on long-short term memory (LSTM) to predict the ODO level from historical measurements, such that if the new measurement differs a lot from the prediction, we could consider that anomaly appears in the lake. We consider that the data from different locations represent different features of the water of the lake. 

In our distributed scheme, each node feeds the time series ODO measurements with window size to be 50 for compression. The first layer is a LSTM with 40 states, then the second is a LSTM layer with 9 states, and the last is a $q$-bits quantization layer. Therefore, each node will send 9 $q$-bits. The base station feeds the 36 inputs to a fully connected layer with 20 outputs, and then a softmax layer with 20 outputs. For the centralized scheme, the nodes transmit all raw measurements to the cloud for data analysis. In this case, the input is 4 time series of ODO data with window size 50 for each time series. We run 10 random trials for both frameworks, and test different quantizations. We average the accuracy results in Table~\ref{table:iWater}. We observe that, the accuracy of our scheme is close to the centralized one, which means that distributing the learning to the nodes locally does not degrade too much in the classification performance. We could also observe that, using 3-bits quantization when the node transmits their local inferences to the base station is good enough. In such a case, each node only transmits $27$ bits for each prediction task.

Based on the evaluation using MNIST and iWater dataset, we observe that our scheme is better than FDML, and our scheme can greatly reduce the data for transmission in the inferring while the degradation in classification is small.

\begin{table*}[t]
	\caption{Comparison of the accuracy achieved by our distributed learning framework and quantizations. The dataset is iWater data set}\label{table:iWater}
	\centering
	\begin{tabular}{|c|l|l|l|l|l|}
		\hline
		Network  & \begin{tabular}[c]{@{}l@{}}Centralized,\\ no quantization\end{tabular} & \begin{tabular}[c]{@{}l@{}}Distributed,\\ no quantization\end{tabular} & \begin{tabular}[c]{@{}l@{}}Distributed, \\ $q=2$\end{tabular} & \begin{tabular}[c]{@{}l@{}}Distributed,\\ $q=3$\end{tabular} & \begin{tabular}[c]{@{}l@{}}Distributed,\\ $q=4$\end{tabular} \\ \hline
		Accuracy & 0.867                                                               & 0.857                                                               & 0.840                                                                     &
		0.845 &		
		0.848                                                                   \\ \hline
	\end{tabular}
\end{table*}

\section{Summary and Future Directions}
This article proposes to model the IoT as an efficient Deep Neural Network. Specifically, we proposed a distributed ML framework that design the compression at the nodes and the inference at the base stations together, such that the nodes just need to transmit little data to the base station. The numerical results showed our framework only losses 2.5\% in inference accuracy while saving 96\% of data transmission, compared to the scheme that sends all raw data. With our framework, one can
develop an extremely efficient IoT systems for monitoring and inference.

There are multiple interesting topics to be further studied to improve the system performance. One is to study the trade-off between communication rate and classification accuracy. Is there a better and systematic way to find the sweet spot? Which layers could greatly save communication while not affecting too much in inference performance? Besides, the data generated by nodes might be re-used in multiple applications. It is possible to have multiple sub neural networks for inference at the base station side, and the loss function for the entire system would take the inference performance of these applications into account. It will be important to design the ML layers and the learning, such that the codes as the output of the nodes can result in good inference performances for multiple monitoring applications.


%



%
%

\ifCLASSOPTIONcaptionsoff
  \newpage
\fi



%
%
%
\bibliographystyle{IEEEbib}
\bibliography{refs} 

\end{document}